\documentclass{article}

\PassOptionsToPackage{numbers, compress}{natbib}

\usepackage[main, final]{neurips_2026}

\usepackage[utf8]{inputenc} 
\usepackage[T1]{fontenc}    
\usepackage{hyperref}       
\usepackage{url}            
\usepackage{booktabs}       
\usepackage{amsfonts}       
\usepackage{nicefrac}       
\usepackage{microtype}      
\usepackage{xcolor}         

\usepackage[pdftex]{graphicx}
\usepackage{amsmath}

\title{$\Psi$-Resilience: Model-Free Feature Importance from 1D Topological Signals}

\author{
  Fabian Galiș \\
  West University of Timișoara \\
  \texttt{fabian.galis@e-uvt.ro} \\
  \And
  Darian Onchiș \\
  West University of Timișoara \\
  \texttt{darian.onchis@e-uvt.ro} \\
  \AND
  Pedro Real Jurado \\
  University of Sevilla \\
  \texttt{real@us.es} \\
}

\begin{document}

\maketitle

\begin{abstract}
We introduce $\Psi$-Resilience, a model-free feature importance method that derives explanations directly from the data itself via 1D topological signals. Our method constructs a class-disagreement landscape by estimating class-conditional densities and taking their pointwise absolute difference along the feature axis. Then, the 0-dimensional persistence of this 1D signal defines a resilience functional that aggregates only those topological features that survive perturbations up to a robustness scale which is set by the user. This gives us a context-robust importance score that is inherently auditable via the underlying 1D landscapes and their persistence. We evaluate our method on both synthetic and real datasets. On synthetic generators with specified ground-truth importance, $\Psi$-Resilience recovers the ranking of features with high fidelity, achieving Spearman rank correlations up to 0.8 and performing competitively with multiple feature importance methods, including SHAP and mutual information. On real datasets with no known ground truth, our technique agrees with these methods, with correlations up to 0.9. These results show that $\Psi$-Resilience is a stable explanation method that enables rigorous, distribution-level auditing of feature importance without relying on a predictive model.
\end{abstract}

\section{Introduction}

Feature importance has become the standard tool for communicating and assessing the behavior of complex predictive models, but despite its frequent usage, there's an ongoing confusion over what exactly is a ``feature importance score'' measuring. Model-based scores such as SHAP values \cite{Lundberg2017} quantify how a particular fitted model changes its predictions when a feature is perturbed, while regulators and domain experts often ask a different question: How important is this feature within our data, regardless of any particular model we may use?

Most works formalise this distinction between model-specific and intrinsic (also called algorithm-agnostic) importance \cite{williamson2023general,saarela2021comparison}. Model-specific measures depend, as the name suggests, on the learning algorithm, but also on its hyperparameters, and even random initialisation, and can disagree dramatically across equally accurate models trained on the same data \cite{saarela2021comparison,parr2024nonparametric}. Intrinsic importance is defined as a functional of the joint distribution of $(X,Y)$, typically via a population-level contrast in predictive performance with versus without a feature, or feature subset \cite{williamson2023general,williamson2021nonparametric}. This yields a property of the data-generating process itself, not of any particular estimator.

Recent work has made a lot of progress on intrinsic importance but there are still large gaps in this regard. Many estimators are basically model-mediated even when the target is conceptually model-free: they require fitting flexible predictive models as nuisance regressors, with final importance estimates sensitive to modelling choices and optimisation instability \cite{williamson2023general}. Another problem is that existing scores based on data only reveal little about the local structure of class separation along a feature, such as whether importance arises from a single large separation region or from many fragile and small pockets of discriminative signal.

\begin{figure}
  \centering
  \includegraphics[width=1\textwidth]{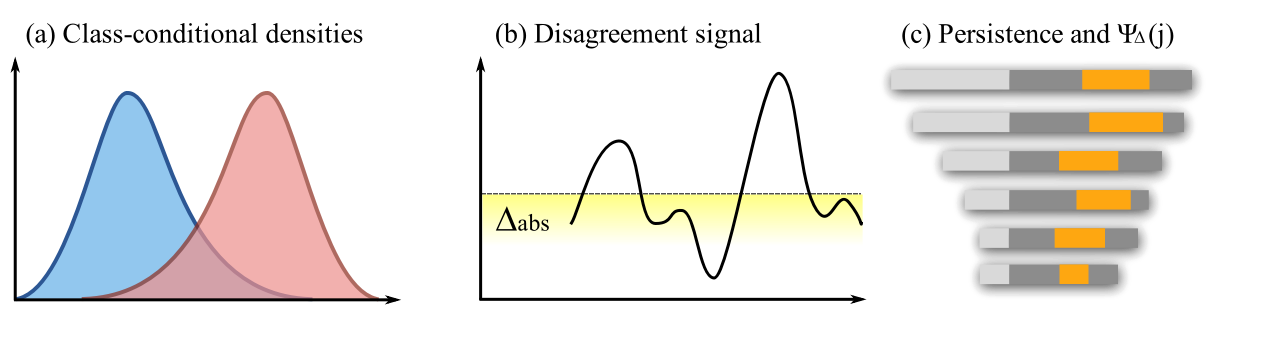}
  \caption{Overview of $\Psi$-Resilience. Starting from class-conditional densities (a), we construct a disagreement signal (b), compute its persistence diagram, and aggregate the $\Delta$-resilient contributions into $\Psi_\Delta(j)$, our final score, for each feature (c).}
  \label{fig:overview}
\end{figure}

In this work, we propose $\Psi$-Resilience: a new approach to model-free global feature importance driven by one-dimensional topological signals. We first construct a class-disagreement landscape for each feature by estimating class-conditional densities and taking their point-wise absolute difference along the feature axis. This gives us a one-dimensional signal whose prominent peaks correspond to the regions in which the feature separates the classes locally. We then build a cubical complex on the negative of this signal and compute zero-dimensional persistence intervals. The resilience functional $\Psi_\Delta$ is the main product of the process - it assigns to each feature the aggregated ``cost'' of removing all topological features (connected components) that survive perturbations up to scale $\Delta$. Intuitively, $\Psi_\Delta$ quantifies how much robust class separation the feature supports at resolution $\Delta$. A general overview of our method can be seen in Figure~\ref{fig:overview}.

We test $\Psi$-Resilience on synthetic benchmarks where ground-truth feature importance is known and compare it to a wide range of intrinsic and model-based global importance measures: mutual information, F-scores, permutation importance, drop-column variable importance \cite{williamson2023general}, SHAP \cite{Lundberg2017}, SAGE \cite{covert2020understanding}, and nonparametric partial-dependence impact \cite{parr2024nonparametric}. The comparison of our method with model-based approaches is intended to move beyond the validation of whether a model behaves like other models with respect to feature importance, to the capturing of the real structure of data and to determine its intrinsic geometry, rather than just fitting to noise. The $\Psi_\Delta$ score obtains competitive alignment with ground-truth rankings while offering stability and auditability properties that are unlike these existing methods. Our results show that $\Psi$-Resilience can serve as a ``topology-for-auditors'' layer that can work in tandem with model-based explainability by exposing the robust separation structure inherent in the data itself.

\section{Related Work}
Much of the current research in explainable AI focuses on \textbf{feature importance for a particular predictive model}. Well-known examples include SHAP \cite{Lundberg2017}, permutation importance (originally developed in the context of random forests), as well as ``leave-one-feature-out'' and ``drop-column'' importance scores. The above methods detail how a certain model $\hat f$ uses the features of the data, either in a local or global context. Yet, as Williamson et al.\ mentioned in the recent literature, these features do not tend to indicate the feature's intrinsic predictive capability. Two models might have the same accuracy, but assign widely divergent importance scores to features, given the same data \cite{williamson2023general,saarela2021comparison,parr2024nonparametric}.

Covert et al.\ propose SAGE, a Shapley-style global measure of importance that combines the marginal contributions of a given feature for predictive accuracy over all possible subsets of features \cite{covert2020understanding}. SAGE is algorithm-agnostic because it is defined in terms of an abstract predictive loss function, but it's instantiated with a chosen model and an imputation strategy, so the resulting scores are still model-dependent. Saarela and Jauhiainen conducted a comparison study demonstrating large variability across various algorithm/model-based measures of importance (such as SHAP, permutation importance, among others) for classification tasks in various models, concluding that interpretations of results for predictive accuracy should be taken cautiously \cite{saarela2021comparison}. The risk of over-interpretation of algorithm/model-based measures of importance has further been noted in recent biomedical research, urging for greater statistical verification of these claims \cite{oka2025beyond,takefuji2025beyond}.

We emphasise in this regard that we are not proposing yet another importance measure that might work well on a particular model, but develop a purely model-independent function that can operate together with the others. The purpose of $\Psi_\Delta$ can be seen as capturing the robust separation information within the data.

Regarding \textbf{intrinsic and algorithm-agnostic variable importance}, Williamson et al.\ defines this notion in more concrete terms by considering a contrast in oracle predictiveness between the set of all available features and a set where a particular set $S$ of features becomes unavailable \cite{williamson2023general,williamson2021nonparametric}. This introduces a set of target quantities that depend solely upon the distribution of $(X,Y)$ and the chosen loss function, as opposed to being sensitive to specifics of learning algorithms. They provide semiparametric efficient estimates using flexible machine learning as nuisance regressors in low and high dimensions.

Building upon this viewpoint, \cite{huang2025factor} establishes that the importance parameters in the Williamson et al.\ method are equivalent to the global sensitivities calculated using the Sobol’ indices. This relationship gives birth to FIRST, a factor importance ranking and selection procedure, specialized in regression, that can be directly approximated from noisy observational data, removing the necessity for the construction of a parametric surrogate. The paper illustrates that the estimation of the total Sobol’ indices can in fact be both statistic-justified and model-agnostic and further supports the idea that the definition of importance should instead be based at the scale of the data-generating mechanism.

These methods all share the common target of specifying importance as a model-free function of the joint distribution. They mainly focus on a loss difference perspective, without revealing the intricate details of the separation between classes along a feature dimension. Instead, $\Psi_\Delta$ is derived from one-dimensional class disagreement signals, making the spatial separation patterns fully inspectable.

There are methods that seek to estimate \textbf{feature importance straight from the data} without fitting a predictive model at all. Parr et al.\ consider feature impact to be a measure of the area under the ideal partial dependence curve and feature importance as a weighted distribution form of this curve. Both of these measures can be expressed as a function of conditional expectations over a conditional joint distribution of $(X,Y)$ and can each be approximated, nonparametrically. Empirically, their values rank well in a comparative manner with other traditional model-based feature selection algorithms \cite{parr2024nonparametric}. Oka et al., as well as Takefuji, argue that correlation, mutual information, and data-only conditions are valuable secondary tools for feature relevance \cite{oka2025beyond,takefuji2025beyond}. $\Psi$-Resilience is closest in spirit to these data-driven approaches, as $\Psi_\Delta$ takes class-conditional probabilities and signals of class disagreement from $(X, Y)$, rather than requiring or relying on predictive models. Our method differs from impact measures derived from partial dependence \cite{parr2024nonparametric} through favoring geometry related to class decomposition in 1D rather than the study of regression-like response curves. In comparison to these methods, ours is accompanied by both importance scores and the robustness parameter $\Delta$.


\section{Methodology}
\label{sec:methodology}

$\Psi_\Delta$ can be viewed as an intrinsically model-free importance functional that operates on one-dimensional density-difference signals encoding local class separation, incorporates an explicit robustness scale via resilience homology, and yields importance scores that are by construction accompanied by interpretable 1D explanation signals.

Let $\mathcal{D} = \{(x_i, y_i)\}_{i=1}^n$ be a supervised classification
dataset with $x_i \in \mathbb{R}^d$ and $y_i \in \{1,\dots,K\}$.
We denote by $x_{i,j}$ the value of feature $j \in \{1,\dots,d\}$ for sample $i$,
and by
\[
\mathcal{D}_{c,j} = \{x_{i,j} : y_i = c\}, \quad c \in \{1,\dots,K\},
\]
the class-conditional sample for feature $j$.

Our goal is to assign each feature $j$ a scalar importance score
$\Psi_\Delta(j)$ that depends only on the joint distribution of $(X,Y)$, and
which quantifies how much robust class-separation structure this
feature supports at a user-chosen robustness scale $\Delta > 0$.

\subsection{Class-Disagreement Signals via KDE}

For each feature $j$, we work with a one-dimensional grid
\[
t_k = a_j + \frac{k}{m-1} (b_j - a_j), \quad k = 0,\dots,m-1,
\]
where $a_j = \min_i x_{i,j}$, $b_j = \max_i x_{i,j}$, and $m \in \mathbb{N}$
is the grid size.

Estimating for each class $c \in \{1,\dots,K\}$ the class-conditional density
of feature $j$ using a univariate kernel density estimator,
\begin{equation}
    \hat{p}_{j,c}(t)
    = \frac{1}{|\mathcal{D}_{c,j}|\, h_j}
      \sum_{x_{i,j} \in \mathcal{D}_{c,j}} 
      K\!\left(\frac{t - x_{i,j}}{h_j}\right),
    \label{eq:kde}
\end{equation}
where $K$ is a fixed kernel (Gaussian in our implementation) and $h_j > 0$ is a
bandwidth parameter. We use Scott’s rule
$h_j = \widehat{\sigma}_j \, n^{-1/5}$, where $\widehat{\sigma}_j$ is the
empirical standard deviation of $x_{i,j}$, but any consistent rule could be
substituted.\\

When $K=2$, let the two classes be $c_0$ and $c_1$. The
disagreement signal for feature $j$ is
\begin{equation}
    f_j(t) = \bigl| \hat{p}_{j,c_1}(t) - \hat{p}_{j,c_0}(t) \bigr|,
    \label{eq:binary-disagreement}
\end{equation}
and we denote by $\mathbf{f}_j = (f_j(t_0),\dots,f_j(t_{m-1}))$ its discrete
vector representation.

When $K>2$, we define pairwise signals.

For each unordered pair $(c_1,c_2)$ with $c_1 < c_2$, we compute
\begin{equation}
    f_{j}^{(c_1,c_2)}(t)
    = \bigl| \hat{p}_{j,c_1}(t) - \hat{p}_{j,c_2}(t) \bigr|.
    \label{eq:pairwise-disagreement}
\end{equation}

In the end, we obtain one or more one-dimensional scalar fields on the
grid $\{t_k\}$, which we treat as topological signals encoding local class
separation along feature $j$.

\subsection{Resilience Homology on 1D Signals}

Given a discrete one-dimensional signal
$\mathbf{f} = (f(t_0),\dots,f(t_{m-1}))$, we construct a cubical complex
$\mathcal{C}$ on these $m$ grid points and the $m-1$ intervals
$[t_k,t_{k+1}]$ between them. We then consider the scalar field $-f$ on
$\mathcal{C}$, assigning each vertex the value $-f(t_k)$ and extending to
edges via the standard upper-star filtration \cite{edelsbrunner2010computational}.

We then form the sublevel-set filtration of $-f$,
\begin{equation}
    \mathcal{C}^\alpha = \{\sigma \in \mathcal{C} : (-f)(x) \le \alpha
                         \text{ for all } x \in \sigma\}, \quad \alpha \in \mathbb{R},
\end{equation}
and compute the $0$-dimensional persistent homology of this filtration, obtaining
a multiset of birth–death pairs
\begin{equation}
    \mathrm{Dgm}_0(f) = \{(b_\ell, d_\ell)\}_{\ell=1}^{L},
    \quad b_\ell < d_\ell,
\end{equation}
where $L$ is the number of finite bars. Each bar corresponds to a connected
component of the sublevel sets that is born at level $b_\ell$ and merges at
level $d_\ell$.

We define the base resilience of bar $\ell$ as half its persistence:
\begin{equation}
    r_\ell = \frac{d_\ell - b_\ell}{2}.
\end{equation}
$r_\ell$ approximates the minimal $L_\infty$ perturbation of $f$
required to eliminate that topological feature in the ideal continuous setting.

Let $\Delta_\mathrm{abs} \ge 0$ be a robustness scale in the same units as $f$.
We define the $\Delta$-resilient contribution of bar $\ell$ as
\begin{equation}
    r_\ell^{(\Delta_\mathrm{abs})}
    = \max\{ r_\ell - \Delta_\mathrm{abs}, 0 \}.
\end{equation}
Bars with $r_\ell \le \Delta_\mathrm{abs}$ are considered too fragile to be
relevant at scale $\Delta_\mathrm{abs}$ and contribute zero; longer bars
contribute linearly beyond this threshold.\\

Then, we aggregate the $\Delta$-resilient contributions across bars using a chosen aggregation operator $\mathsf{Agg}_\mathrm{bar} \in \{\mathrm{sum},\mathrm{mean},\mathrm{max}\}$:
\begin{equation}
    \Psi_\Delta(f)
    = \mathsf{Agg}_\mathrm{bar}
      \bigl( \{ r_\ell^{(\Delta_\mathrm{abs})} \}_{\ell=1}^{L} \bigr),
    \label{eq:psi_single_signal}
\end{equation}
where $\Delta_\mathrm{abs}$ is either a user-specified absolute threshold or,
as in our implementation, derived from a relative threshold $\Delta > 0$ via
\begin{equation}
    \Delta_\mathrm{abs}
    = \Delta \cdot \bigl(\max_k f(t_k) - \min_k f(t_k)\bigr),
    \label{eq:delta_fraction}
\end{equation}
when a relative robustness scale is desired.

In our experiments we primarily use $\mathsf{Agg}_\mathrm{bar}=\mathrm{max}$,
so that $\Psi_\Delta(f)$ captures the resilience of the single most robust
separation mode in the signal.\\

Finally, the feature-level importance score $\Psi_\Delta(j)$ is defined by aggregating $\Psi_\Delta(f)$ over one or more disagreement signals associated with feature $j$.

When $K=2$, there is a single disagreement signal $f_j$ given by
\eqref{eq:binary-disagreement}, and we set $\Psi_\Delta(j) = \Psi_\Delta(f_j)$.

When $K>2$ and we operate in pairwise mode, we have one disagreement signal
$f_{j}^{(c_1,c_2)}$ per unordered class pair $(c_1,c_2)$. We compute
$\Psi_\Delta(f_{j}^{(c_1,c_2)})$ for each pair and aggregate the score across all pairs.

\section{Experiments}

For reproducibility, all experiments were implemented in Python, utilizing scikit-learn for basic modeling \cite{scikit-learn}, the SHAP library \cite{Lundberg2017} for attributions, and GUDHI \cite{gudhi:CubicalComplex} for computing cubical complexes. The experiments were run in a cloud configuration hosted by Google Colab \cite{colab} (T4 GPU). In terms of hyperparameters, unless otherwise stated, we set the grid size to $m=300$, the robustness parameter to $\Delta=0.03$ (interpreted via \eqref{eq:delta_fraction}), $\mathsf{Agg}_\mathrm{bar}=\mathrm{max}$, and $\mathsf{Agg}_\mathrm{pair}=\mathrm{sum}$. Given fixed $(X,Y)$ and hyperparameters, the procedure is deterministic. Any randomness in our experiments arised only from the synthetic data generator and from external baselines, not from $\Psi$-Resilience itself.

We evaluated the method on both real and synthetic datasets:
\begin{enumerate}
    \item \textbf{Wisconsin Breast Cancer}, a modestly-sized (569 instances with 30 features) medical dataset for binary classification \cite{breast_cancer_wisconsin_(diagnostic)_17};
    \item \textbf{Diabetes 130-US}, a medical dataset with 101k instances and 47 features \cite{strack2014impact};
    \item \textbf{Covertype}, a classification of pixels into 7 forest cover types based on 54 attributes, containing 58k instances \cite{blackard1999comparative};
    \item \textbf{Spambase} for binary spam classification, consisting of 4601 x 57 numeric word-frequency features \cite{uci_spambase_1999};
    \item \textbf{Jannis}, a dataset of considerable size (83.7k instances with 55 features) built for a machine learning competition \cite{guyon2015design};
    \item \textbf{PhishingWebsites}, consisting of 11k instances with 68 features \cite{mohammad2012assessment};
    \item \textbf{HIGGS}, a physics dataset with 1 million instances of 29 features for distinguishing between a signal process which produces Higgs bosons and a background process which does not \cite{baldi2014exotic};
    \item A \textbf{synthetic dataset} where true feature importance is known by design. The generation of this dataset is described in the subsequent section.
\end{enumerate}

To evaluate scalability, We included both small-scale and large-scale datasets, such as the 1 million instance HIGGS dataset.
The synthetic dataset enables the validation of our scores against the ground truth, while the real datasets demonstrates the approach on real-world data where ground truth is unavailable, but where comparisons with other feature importance methods can be made.

\subsection{Synthetic Data Generation}

We built a synthetic generator to obtain multiclass datasets where there is a known ground-truth importance along the feature dimensions. For each of these setups, we draw
$X \in \mathbb{R}^{n \times d}$ from a correlated latent Gaussian
$Z \sim \mathcal{N}(0,\Sigma)$, where $\Sigma$ is either Toeplitz,
equicorrelated, or a random symmetric positive definite matrix. A Gaussian
copula step maps $Z$ to uniform scores and then to user-specified marginals
(e.g., Gaussian, lognormal, uniform, Laplace, exponential), yielding the final
design matrix $X$.

We select $s$ features as informative and endow each class $k \in \{1,\dots,K\}$
with its own linear coefficients $w_{k,j}$ on this subset, with
$w_{k,j} \sim \mathrm{Unif}(a,b)\cdot\{\pm1\}$. The class-specific linear
signals are $s_k = X w_k$, optionally augmented by nonlinear transforms
(square, $\sin$, $\tanh$, \texttt{ReLU}, \texttt{abs}) applied to a fraction of
the informative features and a small number of pairwise interaction terms
(centered products). To control difficulty, each signal $s_k$ is rescaled to a
target signal-to-noise ratio (SNR) and perturbed by Gaussian noise,
\[
\ell_{i,k}
= \mathrm{snr} \cdot \frac{s_{i,k}}{\mathrm{SD}(s_{\cdot,k})} + \varepsilon_{i,k},
\qquad \varepsilon_{i,k} \sim \mathcal{N}(0,1),
\]
and class-specific intercepts $b_k$ are calibrated so that the empirical class
frequencies approximately match a desired probability vector $\pi$ under the
softmax model $p(y_i = k \mid x_i) = \mathrm{softmax}_k(\ell_i + b)$. Labels
are sampled as $y_i \sim \mathrm{Categorical}(p(y_i = \cdot \mid x_i))$.

The global importance of feature $j$ in ground truth is defined as the
$\ell_2$-norm of its class-specific linear coefficients,
$\|w_{\cdot,j}\|_2 = \bigl(\sum_{k=1}^K w_{k,j}^2\bigr)^{1/2}$; the nonlinear weights are used only to enhance realism and to add difficulty levels, not to define the target ranking. The irrelevant features have the same correlation structure as the informative ones, so spurious associations and masking effects arise naturally.

In all our experiments involving this data, we used multiple configurations:
\begin{enumerate}
    \item \textbf{Default}: $n=1000$, $d=30$, $K=3$, $s=10$ informative features, $\mathrm{SNR}=1.5$, Toeplitz or random SPD covariance, Gaussian marginals, no nonlinearity.
    \item \textbf{Varying SNR and classes}: $K \in \{3,5\}$, $\mathrm{SNR} \in \{0.8,1.5,2.0\}$, Toeplitz covariance with $\rho \in \{0.2,0.6\}$.
    \item \textbf{Non-Gaussian + nonlinear}: mixed marginals (Gaussian, lognormal, Laplace, uniform, exponential), nonlinear effects on 50\% of informative features with a few pairwise interactions.
\end{enumerate}

These settings allowed the performance of $\Psi$-Resilience and baseline approaches to be effectively evaluated for varying levels of class complexity, correlation, noise, and marginal structure. The experiments show that the method is robust and resistant against such variations.

\subsection{Experimental Setup}

We now describe the experimental process used to compare $\Psi$-Resilience to various feature importance methods. All methods and generators were reimplemented in Python to the best of our knowledge according to their respective articles using standard libraries.

On each synthetic dataset configuration, we computed global feature importance scores for all features using the following methods:

\begin{itemize}
    \item \textbf{$\Psi$-Resilience ($\Psi_\Delta$).} Our method, with a uniform grid of $m=300$ evaluation points per feature and with $\Delta=0.03$ (taken as a fraction of the signal range).

    \item \textbf{Mutual information (MI).} The univariate mutual information
    $I(X_j;Y)$ estimated by the standard nonparametric estimator for
    classification implemented in scikit-learn, which has been proposed as a model-independent importance measure in a variety of contexts \cite{oka2025beyond,takefuji2025beyond}.

    \item \textbf{ANOVA F-score.} The one-way F-statistic $F_j$ of feature $j$
    versus the class label $Y$. Although based on a parametric mean-shift model, this is just a standard data-only baseline for filtering and ranking features.

    \item \textbf{SHAP.} Global SHAP importance was computed from an
    XGBoost classifier \cite{Chen2016} trained on $(X,Y)$ for a maximum
    depth of $4$. We used the TreeSHAP implementation \cite{Lundberg2017} to obtain the SHAP values for each sample and class and aggregated them by taking the mean absolute value across all samples and classes, yielding a single score per feature.

    \item \textbf{Permutation importance.} We computed Breiman-style permutation importance on the same trained XGBoost model by repeatedly permuting each feature $X_j$ and the negative log-likelihood's reduction (i.e., the increase in multiclass log-loss) \cite{breiman2001random}.
    The score for feature $j$ is the mean change in score over 20 permutations.

    \item \textbf{Drop-column VI.} Following the framework for algorithm-agnostic variable importance \cite{williamson2023general},
    we calculated, for each feature $j$, the difference in empirical risk between the full model trained on all features and a reduced model trained on $X$ with column $j$ removed. Risk was measured as multiclass log-loss on the same training data; the resulting importance is
    $\mathrm{VI}_{\mathrm{drop}}(j) =
      \hat{R}_{-j} - \hat{R}_{\mathrm{full}}$.

    \item \textbf{Parr partial-dependence impact.} We implemented a model-based approximation to the nonparametric feature impact of Parr et al.\
    \cite{parr2024nonparametric}. For each feature $j$, we computed a
    model-based partial dependence profile by overwriting $X_{:,j}$ with grid
    values and averaging the predicted probability of the true class over
    samples. The impact score is the mean absolute deviation of this profile
    from its baseline value (the average predicted probability before
    intervention). This yields a scalar impact measure per feature.

\end{itemize}

The methods mentioned above may be classified into purely data-based, model-free scores ($\Psi_\Delta$, MI, F-score) and model-based, algorithm-agnostic scores (SHAP, Permutation, Drop-column, Parr impact, and SAGE). As far as publicly documented implementations of FIRST go, they are focused on regression and binary classification, and we couldn't integrate this model for the multiclass tasks of our pipeline.

\subsection{Results}

\begin{table}[t]
  \centering
  \small

  \begin{minipage}[t]{0.60\textwidth}
    \centering
    \caption{Spearman correlation \(\rho\) with ground-truth feature importance on synthetic benchmarks for different numbers of classes \(K>2\). Results use default generator hyperparameters and are averaged over 5 random seeds, with a standard deviation less than \(10^{-2}\).}
    \label{tab:synthetic-correlations}
    \begin{tabular}{@{}lcccc@{}}
      \toprule
      \textbf{Method} & \textbf{K3} & \textbf{K4} & \textbf{K5} & \textbf{K6} \\
      \midrule
      \(\Psi\)-Resilience      & 0.748 & 0.759 & 0.818 & 0.806 \\
      \midrule
      Mutual Information       & 0.461 & 0.502 & 0.482 & 0.718 \\
      ANOVA F-Score            & 0.737 & 0.746 & 0.821 & 0.806 \\
      SHAP                     & 0.819 & 0.767 & 0.811 & 0.808 \\
      Permutation importance   & 0.810 & 0.692 & 0.803 & 0.811 \\
      Drop-column              & 0.806 & 0.669 & 0.514 & 0.572 \\
      Partial-dependence       & 0.811 & 0.677 & 0.792 & 0.810 \\
      \bottomrule
    \end{tabular}
  \end{minipage}
  \hfill
  \begin{minipage}[t]{0.36\textwidth}
    \centering
    \caption{Spearman correlation \(\rho\) of \(\Psi\)-Resilience with the SHAP scores on the real-world benchmark datasets. The values are deterministic, per-dataset.}
    \label{tab:realdata-correlations}
    \begin{tabular}{@{}lc@{}}
      \toprule
      \textbf{Dataset} & \(\rho\) \\
      \midrule
      Wisconsin Breast Cancer & 0.811 \\
      Diabetes 130-US         & 0.964 \\
      Covertype               & 0.904 \\
      Spambase                & 0.737 \\
      Jannis                  & 0.675 \\
      PhishingWebsites        & 0.645 \\
      HIGGS                   & 0.690 \\
      \bottomrule
    \end{tabular}
  \end{minipage}
\end{table}

For every dataset-method configuration, we computed a vector of feature importance scores. We measured the agreement between the models and the ground truth using Spearman rank correlation \cite{spearman1961proof} calculated over the features. The value of this coefficient doesn't change under monotone transformations of the scores and highlights the consistency of the rankings.

To complement the correlation comparisons, we also measured the runtime for each method and dataset configuration. For each, we measured the elapsed time from the start of the method’s computation to its completion using high-resolution timers. The time differences across all methods remained consistent and proportional relative to each other and the scale of the data. As a reference, in a representative six-class synthetic setting with default generation parameters, $\Psi$-Resilience completed the process in 0.502 seconds on our hardware configuration. In comparison, other model-based methods, such as SHAP (0.956 seconds for XGBoost plus 0.785 seconds), Permutation Importance (0.956 seconds for XGBoost plus 5.784 seconds), Drop-column (33.3 seconds), and PD-based impact (6.324 seconds), experienced high processing time due to the complexity of the underlying models.

Our experimental outcome for synthetic data (Table~\ref{tab:synthetic-correlations}) and real-world data (Table~\ref{tab:realdata-correlations}) verifies our main points. $\Psi$-Resilience is accurate as a global feature-importance measure, even when compared to model-based benchmarks.\\

On \textbf{synthetic data}, SHAP and $\Psi_\Delta$ both performed well across the configurations. SHAP obtained the best correlation with the ground truth importance in several settings ($\rho=0.819$ for $K=3$ and $\rho=0.811$ for $K=5$), which demonstrated its effectiveness. Nevertheless, $\Psi_\Delta$ always fell in the same performance category as SHAP, permutation importance, and Parr impact, with correlations between $0.748$ and $0.818$. $\Psi_\Delta$ also followed the ANOVA F-score, which also captures univariate signal strength, but is more sensitive to linear mean-shift assumptions and lacks robustness tuning.

Regarding the methods based purely on data, mutual information performed well in certain settings but still remained behind $\Psi_\Delta$ and SHAP, especially for larger $K$ or complex correlation structures. Despite being algorithm-agnostic, Drop-column degraded substantially in some configurations from what we have seen (e.g., $\rho=0.514$ for $K=5$), likely due to the need to refit models on reduced feature sets and the resulting instability in high-dimensional, correlated designs.

The synthetic experiments above have shown that the proposed $\Psi_\Delta$ method estimates the ground truth with similar success to advanced model-based approaches at a lower computational expense.\\

\begin{figure}[t]
  \centering

  \begin{minipage}[t]{0.49\textwidth}
    \centering
    \includegraphics[width=\linewidth]{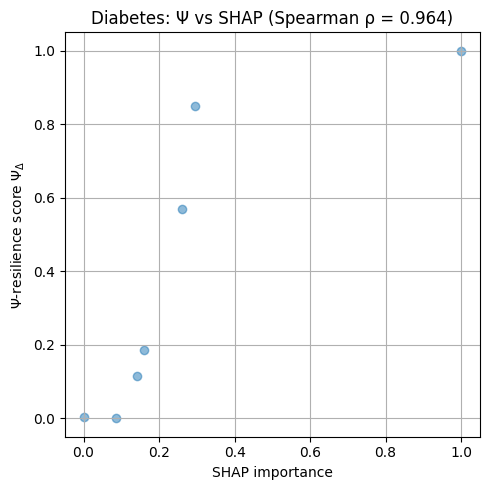}
    \caption{Correlation of \(\Psi\)-Resilience and SHAP scores for the
    Diabetes dataset, scaled to \([0,1]\).}
    \label{fig:diabetes}
  \end{minipage}
  \hfill
  \begin{minipage}[t]{0.49\textwidth}
    \centering
    \includegraphics[width=\linewidth]{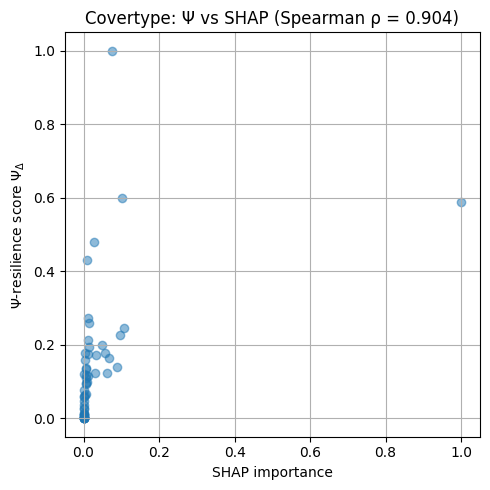}
    \caption{Correlation of \(\Psi\)-Resilience and SHAP scores for the
    Covertype dataset, scaled to \([0,1]\).}
    \label{fig:covertype}
  \end{minipage}

\end{figure}

To examine the relation to $\Psi_\Delta$ in the context of \textbf{real-world data} where the ground truth is not available, we calculated the Spearman rank correlation coefficient between the ranking of SHAP values and $\Psi_\Delta$ features (Table~\ref{tab:realdata-correlations}). These correlations are positive and in many cases high, for instance $\rho=0.964$ for the Diabetes 130-US dataset (as seen in Figure~\ref{fig:diabetes}) and $\rho=0.904$ for the Covertype dataset (as seen in Figure~\ref{fig:covertype}). Even on other challenging, high-dimensional or noisy datasets, the agreement remained moderate to strong ($\rho$ in the range $0.645$--$0.811$). This pattern is encouraging, since in the regimes in which SHAP is trusted to reproduce meaningful model behaviour, $\Psi_\Delta$ is in agreement while being computed without ever fitting a predictive model.\\

In summary, the above experiments demonstrated that $\Psi$-Resilience is able to compete in ranking accuracy on synthetic benchmarks with known ground truth, as well as maintaining consistent agreement with SHAP on real data, while offering distinctive advantages: model independence, robustness control and auditability. All of these properties illustrate that $\Psi_\Delta$ would be a useful addition to the toolbox of global feature-importance methods, especially in scenarios where data-grounded explanations are needed.

\section{Conclusion}
We have introduced $\Psi$-Resilience, a new framework for model-free global feature importance based on one-dimensional topological signals extracted from class-conditional densities. The resulting importance scores are purely data-driven, do not require any predictive model, explicitly parameterised by robustness, and come with a natural visual audit trail via 1D signals and persistence.
On synthetic benchmarks of known ground-truth importance, $\Psi_\Delta$ achieves competitive Spearman rank correlations with the true feature importance, usually falling in the same performance band as SHAP, permutation importance, and partial-dependence impact, and outperforming or matching baselines based only on data, like mutual information or ANOVA F-scores. On real datasets, for which no ground truth is available, $\Psi_\Delta$
shows strong consistency with SHAP-based rankings across a wide collection of classification tasks, suggesting that the captured robust separation patterns are closely aligned with how strong tree-based models take advantage of.

$\Psi$-Resilience can compete with, but also complements existing methods for intrinsic and algorithm-agnostic variable importance and model-based global attribution methods. While loss-based intrinsic importance quantifies how predictive performance changes when features are removed or masked, $\Psi_\Delta$ quantifies how much robust one-dimensional class separation is present along each feature axis, at a specified resolution $\Delta$.

We mention several limitations of our work. First, $\Psi_\Delta$ is fundamentally univariate: it operates on one feature at a time and captures the marginal class-separation structure along that axis. While this is shared by many screening and ranking approaches (MI, F-scores, univariate PD-based impact), it is indicative that purely interaction-driven importance will not be fully captured.
In principle, one could extend the resilience homology to the multivariate disagreement fields of disagreement, but this would introduce major computational and statistical challenges.

Second, we considered a kernel density estimation with a fixed bandwidth rule. Although the method is well-understood and practical in one dimension, the resulting disagreement signals and persistence diagrams are sensitive to bandwidth choice, sample size, and class imbalance. We have adopted a set of standard heuristics (Scott's rule, fixed grids) for robustness and simplicity, but a more systematic treatment of bandwidth selection and uncertainty quantification for $\Psi_\Delta$ remains to be developed.

Finally, we have focused on classification tasks. Adaptation to regression and structured outputs are not trivial.

In summary, we believe that $\Psi$-Resilience is a good example of how tools from topological data analysis can be adapted to obtain simple, robust and auditable feature-importance measures that speak directly to the structure of the data, and that such model-free perspectives play an increasingly important role in building trustworthy and interpretable machine learning systems.



\medskip

\bibliographystyle{plainnat}
\bibliography{neurips_2026}







\end{document}